\documentclass[sigconf]{acmart}
\usepackage{wasysym}
\AtBeginDocument{%
  }

\acmYear{2026}

\usepackage{amsmath,amsfonts}
\usepackage{graphicx}
\usepackage{textcomp}
\usepackage{xcolor}
\usepackage [english]{babel}
\usepackage [autostyle, english = american]{csquotes}
\MakeOuterQuote{"}
\usepackage{enumitem}
\usepackage{graphicx}
\usepackage{float}
\usepackage{dsfont}
\usepackage{graphics} 
\usepackage{epsfig} 
\usepackage{tabularx} 
\usepackage{multicol}
\usepackage{multirow}
\usepackage{tabularx}
\usepackage{pdflscape}
\usepackage[noindentafter]{titlesec}
\usepackage{bbm}
\usepackage{balance} 

\usepackage{wasysym}
\usepackage{pifont}
\usepackage{afterpage}
\usepackage{capt-of}
\usepackage{bbm}
\usepackage{rotating}
\usepackage{lscape}
\usepackage{booktabs}
\usepackage{hyperref}
\usepackage{makecell}
\usepackage{pgfplots}
\pgfplotsset{width=8cm,compat=1.9}
\graphicspath{ {images/} }
\usepackage{braket} 
\usepackage{algorithm}
\usepackage{amsthm}
\usepackage{xcolor}
\usepackage{mathtools}

\usepackage[noend]{algpseudocode}

\usepackage{titlesec}
\titlespacing{\section}{0pt}{\baselineskip}{1pt}  
\titlespacing{\subsection}{0pt}{\baselineskip}{1pt}

\begin{document}
\urlstyle{tt}
\title{SetAD: Semi-Supervised Anomaly Learning in Contextual Sets}

\author{Jianling Gao}
\email{jianlingg@buaa.edu.cn}
\affiliation{%
  \institution{SKLCCSE, Beihang University}
  \city{Beijing}
  \country{China}
}

\author{Chongyang Tao}
\email{chongyang@buaa.edu.cn}
\affiliation{%
  \institution{SKLCCSE, Beihang University}
  \city{Beijing}
  \country{China}
}

\author{Xuelian Lin}
\email{linxl@buaa.edu.cn}
\affiliation{%
  \institution{SKLCCSE, Beihang University}
  \city{Beijing}
  \country{China}
}

\author{Junfeng Liu}
\email{liujf@pcl.ac.cn }
\orcid{0009-0007-9715-6034}
\affiliation{%
  \institution{Pengcheng Laboratory}
  \city{ShenZhen}
  \country{China}
}

\author{Shuai Ma}
\email{mashuai@buaa.edu.cn}
\affiliation{%
  \institution{SKLCCSE, Beihang University}
  \city{Beijing}
  \country{China}
}

\renewcommand{\shortauthors}{}

\begin{abstract}
    Semi-supervised anomaly detection (AD) has shown great promise by effectively leveraging limited labeled data. However, existing methods are typically structured around scoring individual points or simple pairs. Such {point- or pair-centric} view not only overlooks the contextual nature of anomalies, which are defined by their deviation from a collective group, but also fails to exploit the rich supervisory signals that can be generated from the combinatorial composition of sets. Consequently, such models struggle to exploit the high-order interactions within the data, which are critical for learning discriminative representations.
    To address these limitations, we propose SetAD, a novel framework that reframes semi-supervised AD as a \underline{Set}-level \underline{A}nomaly \underline{D}etection task. SetAD employs an attention-based set encoder trained via a graded learning objective, where the model learns to quantify the degree of anomalousness within an entire set. This approach directly models the complex group-level interactions that define anomalies. Furthermore, to enhance robustness and score calibration, we propose a context-calibrated anomaly scoring mechanism, which assesses a point's anomaly score by aggregating its normalized deviations from peer behavior across multiple, diverse contextual sets.
    Extensive experiments on 10 real-world datasets demonstrate that SetAD significantly outperforms state-of-the-art models. Notably, we show that our model's performance consistently improves with increasing set size, providing strong empirical support for the set-based formulation of anomaly detection. 
\end{abstract}




\maketitle

\section{Introduction}
Anomaly detection, the task of identifying data instances that deviate from a norm, is a critical problem with far-reaching applications across diverse domains, including cybersecurity~\cite{garcia2009anomaly}, financial fraud prevention~\cite{dal2017credit, liu2021pick}, industrial monitoring~\cite{liu2024deep}, and medical diagnosis~\cite{henson2021anomaly, fernando2021deep}. The core challenge lies in computationally defining and capturing this "deviation." As articulated in early foundational work, an outlier is an observation that "appears to deviate markedly from other members of the sample in which it occurs"~\cite{grubbs1969procedures}. This definition establishes a fundamental principle: anomalousness is not an intrinsic property of a point, but a relative concept defined by its relationship to a collective context. Recently, the rapid growth of data complexity and volume, coupled with growing demand for automated and robust anomaly detection solutions, has spurred significant research in this area, particularly leveraging deep learning to automatically learn representations that distinguish normal patterns from anomalous ones.

Among different learning settings, the semi-supervised anomaly detection (SSAD) approach has gained significant traction~\cite{ruff2018deep, akcay2019ganomaly, zhou2021feature, pang2023deep, stradiotti2024semi, lu2024robust, xiao2025semi}. By leveraging a small number of labeled anomalies alongside a large pool of unlabeled data, semi-supervised methods offer a practical and effective compromise for real-world scenarios where labeled data is scarce but not entirely absent. While existing deep SSAD methods have achieved considerable success, a closer examination reveals that many of them do not fully embrace this foundational principle of context-dependency. A significant portion of these approaches still evaluates the anomalousness of a data point in a highly localized or isolated manner. For instance, some methods focus on learning a scoring function that maps a single point to an anomaly score, effectively treating each instance independently without explicitly modeling its relationship to its peers~\cite{pang2019deep, zhou2021feature}. This formulation inherently limits the diversity of supervision, as each labeled anomaly contributes to only a narrow slice of the training signal. Other works have advanced to a pairwise formulation, learning to discriminate between normal-normal and normal-anomaly pairs ~\cite{pang2023deep} or measuring the distance of each point to a learned data center ~\cite{ruff2018deep}. Although these pairwise comparisons introduce a basic form of context, a single reference point or a center may provide a narrow and often insufficient view for identifying complex anomalies. 
{While a few methods, such as REPEN~\cite{pang2018learning}, incorporate multiple points by independently encoding them and computing the k-NN distance of a test point to this reference set, their approach remains inherently point-centric. By not explicitly modeling interactions between the test point and its context, it limits expressiveness and the ability to fully leverage available data and discriminate subtle anomalies.}


To address these challenges, we introduce SetAD, a novel framework designed to explicitly operationalize the principle of context-dependent anomaly detection. Instead of evaluating individual points or pairs, SetAD reframes anomaly detection as a \emph{set-level graded learning} problem: it learns to quantify the collective degree of anomalousness within a set of points. The core of our framework is a purpose-built interaction-aware set encoder that leverages a self-attention mechanism. This architecture symmetrically models all pairwise and high-order dependencies among the points in a set, allowing it to capture the complex structural distortions caused by anomalies—a capability largely absent in prior work. We train this encoder using a graded regression objective, tasking it to predict the number of known anomalies within dynamically sampled sets. This formulation transforms a small number of labeled anomalies into a rich, combinatorially diverse training signal, significantly amplifying supervision beyond traditional point-wise or pairwise schemes. During inference, we employ a robust \emph{context-calibrated anomaly scoring} mechanism, which provides a stable and calibrated score for any test point by evaluating it against its peers within different contexts. More importantly, our extensive experiment has shown that this holistic, set-based design allows our model's performance to consistently improve with increasing context size, providing strong empirical validation for our central hypothesis.

To sum up, the main contributions of this work are as follows:
\begin{itemize}
    
    

    \item We propose SetAD, a novel framework for semi-supervised anomaly detection,  which reframes the task from point-wise evaluation to quantifying the collective anomalousness of a set. 
    \item  We design a set-level graded learning framework that incorporates an attention-based encoder to model intra-set dependencies and capture the context-sensitive nature of anomalies.
    
    \item We present a context-calibrated anomaly scoring mechanism that 
    yields stable scores by normalizing each test point against its contextual peers, mitigating biases from sampling noise and data contamination.
    \item We conduct comprehensive experiments on 10 real-world benchmark datasets, demonstrating that SetAD significantly outperforms state-of-the-art methods. Furthermore, we provide the empirical evidence that a well-designed set-based model's performance consistently improves with increased context size, validating our core hypothesis.
\end{itemize}

\section{Related Works}
\subsection{Unsupervised Anomaly Detection}
Unsupervised anomaly detection has been extensively studied, driven by the scarcity of labeled data in many real-world applications. These methods operate under the assumption that anomalies are rare and structurally different from the majority of normal data. Classical methods often define anomalies based on statistical or distance-based heuristics. For example, the widely-used Local Outlier Factor (LOF)\cite{breunig2000lof} identifies points in sparse neighborhoods as anomalies by comparing local densities. Isolation Forest (IF)\cite{liu2008isolation} identifies anomalies as points that are more susceptible to isolation through random feature partitioning. One-class classification methods, including One-Class SVM~\cite{scholkopf2001estimating} and its deep variant Deep SVDD~\cite{ruff2018deep} attempt to enclose normal data within a compact boundary and treat deviations as outliers. Distribution-based approaches like ECOD~\cite{li2022ecod} and COPOD~\cite{li2020copod} estimate the distribution of data and identify anomalies via statistical extremities.

Deep learning-based methods further extend these paradigms. Autoencoders~\cite{zhou2017anomaly, chen2017outlier} reconstruct inputs and consider high reconstruction error as indicative of anomalies. Generative Adversarial Networks (GANs)\cite{schlegl2017unsupervised, zenati2018efficient} learn to generate data and use a discriminator to flag data that do not align with learned distributions. Recent advancements have focused on learning the intricate structure of normal data through representation learning. Deep Isolation Forest~\cite{xu2023deep} improves IF by replacing data vector with deep feature representations, which enables better partitioning in complex spaces. DPAD~\cite{fu2024dense} proposes learning dense, locally compact embeddings for normal data, and evaluates anomalies by measuring sparsity in the learned space using kNN. Unlike traditional dimensionality reduction techniques, DPAD explicitly optimizes for compactness rather than reconstruction or visualization quality. These methods reflect a growing trend toward learning representation spaces where anomalies become easier to separate.

Despite these significant advancements, unsupervised methods face a fundamental challenge: without any labeled guidance, they rely on strong, often implicit assumptions about the nature of anomalies (e.g., low density, high reconstruction error). These assumptions may not hold true for all types of anomalies, leading to performance limitations. This limitation highlights the need for supervision to learn more nuanced and context-aware anomaly definitions, which is the focus of semi-supervised approaches.

\begin{figure*}[t]
    \centering
    \vspace{-2mm}
    \includegraphics[width=0.92\textwidth, trim={0 0.6cm 0 0.3cm}, clip]{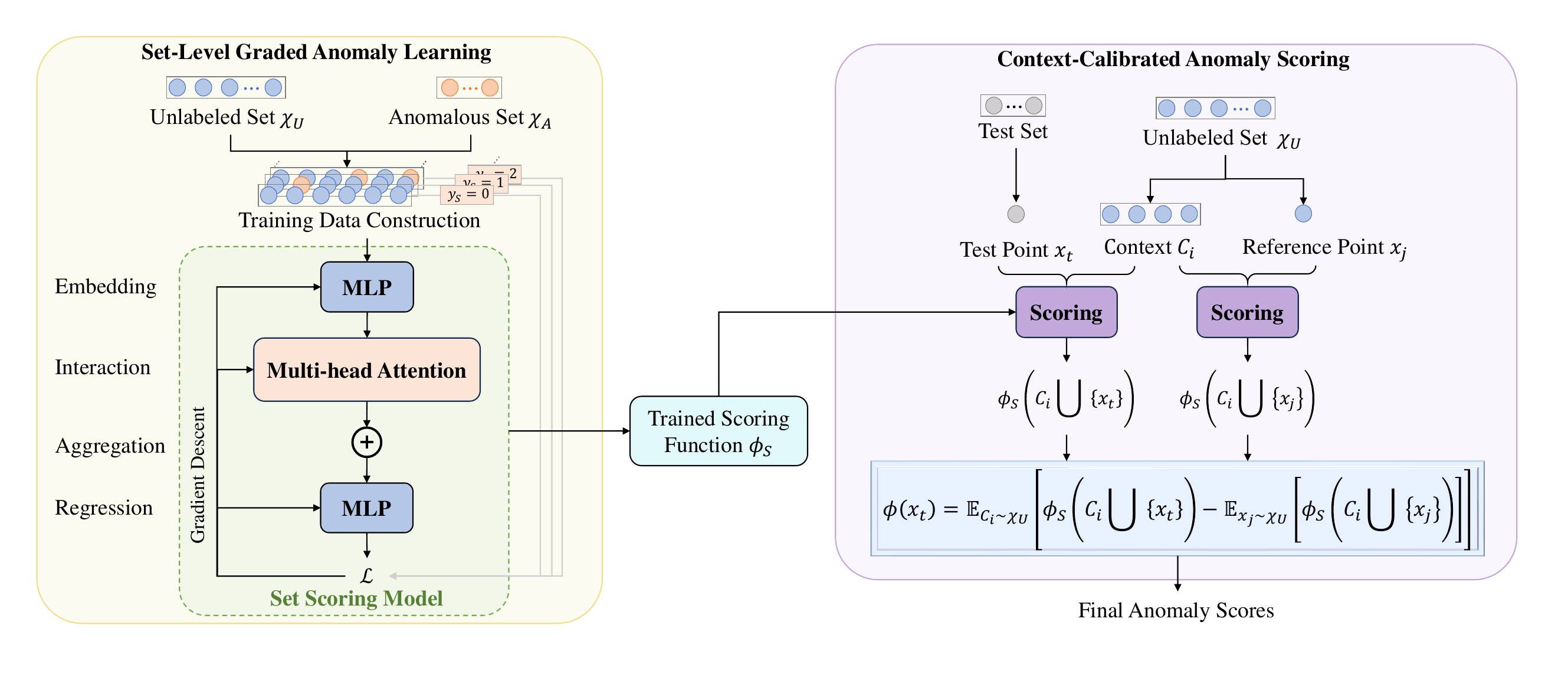}
    \vspace{-5.5ex}
    \caption{The proposed SetAD framework. The framework operates in two phases. (Left) A set scoring model is trained via graded regression to predict the number of anomalies in dynamically sampled sets. (Right) During inference, the trained model scores a test point by normalizing its score within shared contexts against the expected score of reference points.}
    \vspace{-3ex}
    \label{fig1}
\end{figure*}

\subsection{Semi-supervised Anomaly Detection}
The limitations of unsupervised methods have spurred the development of semi-supervised anomaly detection, which aims to improve performance by leveraging a small number of labeled anomalies. A typical way in SSAD is to intergrade supervision signal to guide the learning process in unsupervised method. DevNet\cite{pang2019deep} exemplifies the point-wise approach, training an end-to-end network to directly output an anomaly score for each instance, guided by a prior on the score distribution. Similarly, FEAWAD\cite{zhou2021feature} uses labeled data to guide the feature extraction of an autoencoder, thereby improving its point-wise scoring capability.
GANomaly~\cite{akcay2019ganomaly} takes a generative approach by combining adversarial learning with reconstruction-based detection, using an encoder–decoder–encoder architecture to reconstruct normal data and identify anomalies via reconstruction and latent inconsistencies.
Moving to pairwise relationships, DeepSAD\cite{DeepSAD} extends the unsupervised Deep SVDD by using labeled anomalies to enforce a clearer separation, pushing normal data towards a learned hypersphere center while repelling anomalies. PReNet\cite{pang2023deep} further formalizes this by learning a dedicated relation network to explicitly score the similarity of data pairs. SSIF\cite{stradiotti2024semi} adapts the classic Isolation Forest by using labeled instances to probabilistically guide the tree-building process, biasing splits to more effectively isolate anomalies. REPEN~\cite{pang2018learning} scores a point based on its k-nearest neighbor distance to a reference set of normal samples. While it considers a "set", the underlying mechanism is an aggregation of simple pairwise Euclidean distances, which does not capture the complex, high-order interactions within the combined group of the target point and its context. Other innovative approaches, such as SSRDT~\cite{xiao2025semi}, operate on a higher level by learning a transformation that projects the entire distribution of normal and anomalous data into two distinct, well-separated target distributions in a latent space. Several newer works tackle more specialized challenges in SSAD. For example, TargAD~\cite{lu2024robust} focuses on identifying user-defined or domain-specific anomalies, while CAD~\cite{gao2025semi} proposes a denoising framework tailored for highly contaminated training data. NNG-Mix~\cite{dong2024nng} augments training with synthetic anomalies to enrich supervision, and KDAlign~\cite{zhao2024weakly} leverages external knowledge to improve detection accuracy.

While these methods offer diverse perspectives, most still operate on point-level or pair-level comparisons. This paradigm may have two limitations. First, it fails to model the holistic, internal structure of the context, which is often what truly defines a sophisticated anomaly. An anomaly's nature may not be revealed by its distance to a center or its relationship with a single peer, but by how it disrupts the collective geometry of a group. Second, this paradigm underutilizes the available supervision. By focusing on individual points, it misses the opportunity for rich combinatorial supervision that arises from sets. A handful of labeled anomalies can generate a vast number of unique sets with varying anomaly grades, providing a much denser and more nuanced training signal.
\section{Methodology}

\subsection{Problem Statement}
Let $\mathcal{X}=\{x_1, ..., x_n, x_{n+1}, ..., x_{n+m}\} \subset \mathbb{R}^d$ be a dataset drawn from an unknown data distribution $p(x)$, where $\mathcal{X}_A=\{x_{n+1}, x_{n+2}, ..., x_{n+m}\}$ represents a small set of labeled anomalies, and $\mathcal{X}_U=\{x_1, x_2, ..., x_n\}$ is a large pool of unlabeled data, primarily consisting of normal points. We assume that $m \ll n$. The goal of a semi-supervised anomaly detection model is to learn a scoring function $\phi:\mathbb{R}^d \rightarrow \mathbb{R}$ such that $\phi(x) \propto log \frac{p(x|y=1)}{p(x|y=0)}$, where $y=1$ indicates that the point $x$ is anomalous, and $y=0$ indicates that the point $x$ is normal. The function $\phi(x)$ is designed to assign a higher score to anomalous points, distinguishing them from the majority of normal points.

\subsection{SetAD Overview}
\paragraph{Reframing AD as a Set Scoring Problem.} 
Anomaly is inherently a relative concept, which is usually defined as a data instance that deviates significantly from others. This motivates a context-dependent formulation, in which the anomalousness of a point should be evaluated in relation to other samples. In this work, we propose SetAD, a novel approach that reframes anomaly detection as a \emph{set-level graded learning} task. Instead of scoring individual points in isolation, SetAD learns to measure the degree of anomalousness of a set based on the number of anomalies it contains. By training on sets with varying anomaly counts, the model learns to quantify how the inclusion of anomalous samples distorts the overall structure of the set. This set-based formulation unifies and generalizes point-wise and pairwise approaches, and enables rich combinatorial supervision even with a limited number of labeled anomalies. 

Specifically, the overall architecture of SetAD is illustrated in Fig.\ref{fig1}. The training data is constructed by randomly sampling sets containing varying numbers of labeled anomalies, which encourages the model to learn graded representations across a continuum of anomaly severity.  The scoring function \(\phi(\cdot)\) is implemented via an auxiliary function \(\phi_S(\cdot)\) that evaluates the anomaly level of a set \(S\). An attention-based encoder is first used to aggregate the information of the input set and get the joint embedding of the whole set. This allows the model to capture complex inter-point relationships across the set. This representation is then passed through a regression head to produce a scalar score. By training the model to regress toward the number of anomalies in each set, SetAD benefits from smoother supervision than hard-label classification, and learns to capture fine-grained inconsistency patterns among set elements. During inference, to reduce the impact of randomness and mitigate the influence of contamination in the training data, we apply a \emph{context-controlled normalization strategy} to provide robust and calibrated anomaly scores.

\subsection{Set-Level Graded Anomaly Learning}
This section introduces the core learning mechanism of SetAD, which aims to quantify the degree of anomalousness for a given set of instances. To learn representations that are sensitive to the presence of anomalies, we move beyond existing pointwise or pairwise methods. Our approach learns to regress a continuous score that reflects the anomaly severity of a set. This enables the model to capture subtle inconsistency patterns and leverage richer supervision from limited labeled data through various combinations of data instances. The remainder of this section details the training data construction, the architecture of the set encoder, the training objective, and the overall training procedure.

\subsubsection{Set Construction and Supervision Signal.} 
To train the set-level scoring function \(\phi_S\), we construct synthetic training samples by randomly sampling small sets of data points from the training pool. Specially, each set \(S = \{x_1, x_2, \dots, x_{k}\}\) of size $k$ contains a mix of labeled anomalies from \(\mathcal{X}_A\) and unlabeled samples from \(\mathcal{X}_U\). The supervision signal for each set \(y_S=n_A\) is defined as the number of known anomalies it contains (e.g., 0, 1, or 2). This sampling strategy introduces graded supervision, allowing the model to learn how anomaly severity increases as more anomalous instances are included in the set. 

We acknowledge that unlabeled data $\mathcal{X}_U$ may contain unknown anomalies, leading to imperfect set labels during training. However, our model is designed to tolerate such contamination: the contamination rate is low, mislabeled sets are treated as stochastic noise, and our following graded, set-based training encourages learning of relative anomaly severity. Moreover, inference-time normalization further mitigates residual effects, as detailed in Section~\ref{Inference}.
  

\subsubsection{Set Scoring Model.}
To effectively score a set $S$, the encoder must learn a function that is invariant to the permutation of its elements, as the ordering of points within a set carries no intrinsic meaning. Our architecture achieves this through a sequence of embedding, interaction, aggregation, and regression stages. First, each element \(x_i \in S\) is projected into a shared latent space using an embedding function \(\mathcal{E}: \mathbb{R}^{d} \rightarrow \mathbb{R}^{d_h}\), parameterized by \(\theta_1\):
\begin{equation}
    z_i = \mathcal{E}(x_i; \theta_1), \quad \forall x_i \in S
\end{equation}
where \(d_h < d\) is the hidden dimension of the learned representation. This results in a set of embedding vectors $Z=\{z_1, z_2,...z_k\}$. After that, we employ a multi-head attention mechanism~\cite{vaswani2017attention} to condition the context embedding on the target point. Specifically, the entire set of embeddings $Z$ serves as the Query, Key, and Value for a multi-head attention layers:
\vspace{-0.5em}
\begin{equation}
    Z'=\operatorname{softmax}(\frac{QK^T}{\sqrt{d_h/h}})V
\vspace{-0.5em}
\end{equation}
where $h$ is the number of attention heads, $Q=W_qZ$, $K=W_kZ$,$V=W_vZ$ with learned projection matrices, and $Z' = \{z_1',z_2',...z_k'\}$ is the set of contextualized output embeddings. This step enables the model to capture high-order structural information by creating representations where each point's embedding is informed by its relationship to the entire set.

To produce a single, fixed-size representation for the whole set, we then apply a permutation-invariant sum pooling operation\footnote{We also explored using max pooling, which yielded slightly inferior performance.} to aggregate the contextualized embeddings and accumulate the contributions from each point in the set:
\vspace{-0.5em}
\begin{equation}
z_S = \mathtt{Agg}(Z') = \sum_{i=1}^{s} z'_i
\end{equation}
This aggregation step compiles the learned interactions into a holistic summary vector $z_S$. This vector is subsequently passed through a regression head \(\mathcal{R}: \mathbb{R}^{(k) \times d_h} \rightarrow \mathbb{R}\), parameterized by \(\theta_2\), to produce the final scalar anomaly score:
\begin{equation}
    \phi_S(S) = \mathcal{R}(z_S; \theta_2)
\end{equation}

In our implementation, both the representation function \(\mathcal{E}\) and the regression head \(\mathcal{R}\) are instantiated as multilayer perceptrons (MLPs). {Notably, when the set size is \(k = 2\), the architecture reduces to a pairwise model similar to PReNet~\cite{pang2023deep}, which concatenates the embeddings of a target point with a single randomly sampled instance; when $k=1$, this architecture reduces to a pointwise model similar to DevNet~\cite{pang2019deep}, which takes the embedding of a single point to evaluate its anomalousness.} However, by modeling higher-order interactions among multiple points and combining this with our robust inference strategy, SetAD is able to capture richer contextual cues. As we will demonstrate empirically, detecting anomalies at the set level leads to improved performance over methods that operate on individual points or point pairs.

\setlength{\textfloatsep}{1pt}
\begin{algorithm}[t]
\caption{Set-Level Graded Anomaly Learning}
\label{alg1}
\begin{algorithmic}[1]
    \item[\textbf{Input:}] training set $\mathcal{X} = \{\mathcal{X}_U, \mathcal{X}_A\} \in \mathbb{R}^d$
    \item[\textbf{Output:}] set scoring function $\phi_S : \mathbb{R}^{(k)\times d} \rightarrow \mathbb{R}$ \\
    Randomly initialize model parameters $\Theta =\{\theta_1,\theta_2, W_q, W_k, W_v\}$\\
    \textbf{for} $i=1$ to $n\_epochs$ \textbf{do} \\
    \quad\textbf{for} $j=1$ to $n\_batches$ \textbf{do} \\
    \quad\quad Randomly sample a batch of training pairs $\{(S_i,y_{S_i})\}_{i=1}^B$ \\
    \quad\quad Calculate loss $\mathcal{J} \leftarrow \frac{1}{N} \sum_{i=1}^{N} \left| \phi_S(S_i) - y_{S_i} \right|$\\
    \quad\quad Gradient descent to update model parameters \\
    \quad\textbf{end for}\\
    \textbf{end for} \\
    \textbf{return}  Set scoring function $\phi_S$ parameterized by $\Theta$
\end{algorithmic}
\end{algorithm}
\subsubsection{Training Objective.}
To optimize \(\phi_S\), we use a regression objective that minimizes the mean absolute error (MAE) between the predicted anomaly score and the ground truth label:
\vspace{-0.5em}
\begin{equation}
    \mathcal{J}(\Theta) = \frac{1}{N} \sum_{i=1}^{N} \left| \phi_S(S_i) - y_{S_i} \right|
\end{equation}
where \(N\) is the number of sampled sets in a batch,  the label $y_{S_i}=n_A$ is the number of known anomalies in $S_i$, $\Theta$ denotes all learnable parameters of the model.  By training the model to predict the quantity of anomalies, we provide it with a rich, quantitative supervisory signal. This encourages the model to learn a structured mapping where the output score meaningfully reflects the degree of collective anomalousness. We also experimented with Mean Squared Error (MSE) loss, which yielded comparable performance. 


\subsubsection{Training Procedure.}
The overall training procedure for implementing our Set-Level Graded Anomaly Learning is formally outlined in Algorithm \ref{alg1}. The model is trained iteratively over a series of epochs and mini-batches. In each training step, a new mini-batch of set-label pairs $\{(S_i,y_{S_i})\}_{i=1}^B$ is dynamically generated according to the sampling strategy described previously, where each set $S_i$ is constructed with a known number of anomalies serving as its ground-truth grade, $y_{S_i}$. The set scoring function $\phi_S$ then processes each set to predict its anomaly grade. The Mean Absolute Error between the model's predictions and the true labels is computed across the batch. This loss is then backpropagated to update the model's parameters $\Theta =\{\theta_1,\theta_2, W_q, W_k, W_v\}$ via gradient descent. Through this process, the model learns a scoring function that is explicitly optimized to quantify the degree of anomalousness within a set of instances.

\setlength{\textfloatsep}{6pt}
\setlength{\floatsep}{4pt}
\begin{algorithm}[t]
\caption{Context-Calibrated Anomaly Scoring}
\label{alg2}
\begin{algorithmic}[1]
\item[\textbf{Input:}] Trained set scoring function $\phi_S$, test point $x_t$, normal set $\mathcal{X}_U$, number of contexts $n_C$, number of reference normals $n_r$.
\item[\textbf{Output:}] Anomaly score for the test point $\phi(x_t)$. 
\State Initialize a sum for output scores:$test\_sum \leftarrow 0$ 
\State\textbf{for} $i=1$ to $n_C$ \textbf{do} 
\State \quad Randomly sample a context set $C_i\subset \mathcal{X}_U$ of size $k-1$ 
\State\quad Compute the test score $s_i(x_t) \leftarrow \phi_S(\{x_t\}\cup C_i)$ 
\State\quad Initialize a sum for reference scores:$ref\_sum \leftarrow 0$ 
\State\quad \textbf{for} $j=1$ to $n_r$ \textbf{do} 
\State\quad\quad Randomly sample a reference point $x_j$ from $\mathcal{X}_U$ 
\State\quad\quad Compute the reference score $s(x_j) \leftarrow \phi_S(\{x_j\}\cup C_i)$ 
\State\quad\quad $ref\_sum \leftarrow ref\_sum + s(x_j)$ 
\State\quad \textbf{end for} 
\State\quad Compute the mean reference score:$\bar{s_r} \leftarrow ref\_sum/n_r$ 
\State\quad Compute the normalized test score for this context \hspace*{0.75em} $\bar{s}_i(x_t) \leftarrow s_i(x_t) - \bar{s_r}$ 
\State\quad $test\_sum \leftarrow test\_sum + \bar{s}_i(x_t)$ 
\State\textbf{end for} 
\State\textbf{return} Final score $\phi(x_t) \leftarrow test\_sum/n_C$ 
\end{algorithmic}
\end{algorithm}

\vspace{-0.5em}
\subsection{Context-Calibrated Anomaly Scoring}
\label{Inference}
While our model $\phi_S$ is trained to score sets, the ultimate goal of anomaly detection is to assign a score to an individual data point. A naive approach might be to place a test point in a random set of normal points and use the resulting score directly. However, such a score would lack a meaningful scale and be highly sensitive to the other points in the set, a major issue given that our normal data pool may suffer from contamination by unknown anomalies. For instance, a perfectly normal test point could receive a high score simply because it was paired with a "dirty" set that contains unlabeled anomalies, leading to a false positive. 

To solve this, we introduce a context-calibrated anomaly scoring mechanism,. The core principle is to disentangle a point's intrinsic anomalousness from the properties of the points it is compared against. We achieve this by measuring a point's score relative to a consistent local reference, which we formally define as a context $C$. In our framework, a context is a set of $s-1$ points randomly sampled from the normal data pool $\mathcal{X}_U$. By using this context as a shared background, we can directly compare the effect a test point has versus the effect a typical normal point has.

Our procedure operationalizes this principle through a two-step scoring and normalization process for any given context $C_i$. First, we compute the raw score of the test point $x_t$, within this context, given by $s_i = \phi_S(C_i \cup \{x_t\})$. Second, to establish a dynamic baseline of normalcy specific to $C_i$, we calculate the expected score of a set formed by pairing this same context with a point drawn randomly from the unlabeled pool. The resulting context-normalized score is the deviation of the test score from this peer-based expectation:
\vspace{-0.25em}
\begin{equation}
    \bar{s_i}(x_t) = \phi_S(C_i \cup \{x_t\}) -\mathbb{E}_{x_j \sim \mathcal{X}_U}[\phi_S(C_i \cup \{x_j\})]
\end{equation}
\indent While this normalization corrects for the bias of a context, the score $\bar{s_i}(x_t)$ is still a random variable dependent on the choice of $C_i$. To obtain a stable final score that is independent of any single context's peculiarities, we marginalize out this randomness. This is achieved by computing the expected value of the normalized score over the distribution of all possible contexts that can be sampled from $\mathcal{X}_U$. The final anomaly score $\phi(x_t)$ is therefore defined as:
\begin{equation}
    \phi(x_t) = \mathbb{E}_{C_i \sim \mathcal{X}_U}[\phi_S(C_i \cup \{x_t\}) -\mathbb{E}_{x_j \sim \mathcal{X}_U}[\phi_S(C_i \cup \{x_j\})]]
\end{equation}
\indent This marginalization over the context distribution effectively averages out sampling variance, producing a highly reliable and robust final score. The practical Monte Carlo procedure for approximating these expectations is formalized in Algorithm \ref{alg2}. 
Notably, since the sampling of reference points is independent of the test sample, the expected reference score for each context can be precomputed and reused across all test instances.
Each test sample is evaluated over \(n_C\) sets, each of size \(k\), resulting in \(n_C \times k\) total inputs to the set encoder per test instance. Thanks to the batch-wise set encoding, these computations can be handled within a single forward pass for the attention-base encoder, ensuring efficient inference

\section{Experiments}
\label{exp}

\setlength{\textfloatsep}{12pt}
\begin{table}[t]
\small
    \setlength{\tabcolsep}{2pt}
    \centering
        \caption{Statistics of datasets.}
    \vspace{-2ex}
    \begin{tabular}{c|cccccc}
    \hline
    \textbf{Datasets}                  
    & \textbf{Size} & \textbf{$d$} & \textbf{Anomalies}
     & \textbf{$m$} & \textbf{Category}\\
    \hline
    Cardiotocography & 2114 & 21 & 466 & 18(5\%) & Healthcare\\  

    Mammography & 11183 & 6 & 260 & 10(5\%) & Healthcare\\

    SpamBase & 4207 & 57 & 1679 & 13(1\%) & Document\\   

    Mnist & 7603 & 100 & 700 & 28(5\%) & Image\\  

    Shuttle & 49097 & 9 & 3511 & 28(1\%) & Image\\ 

    Celeba & 202599 & 39 & 4547 & 36(1\%) & Image\\ 
    
    Cover & 286048 & 10& 2747 & 21(1\%) & Botany\\ 

    Campaign & 41118 & 62 & 4640 & 37(1\%) & Finance\\    

    Fraud & 284807 & 29 & 492 & 19(5\%) & Finance\\  

    Census & 299285 & 500 & 18568 & 148(1\%) & Sociology\\  

    \hline
    \end{tabular}
    \vspace{-4mm}
    \label{tab1}
\end{table}

\begin{table*}[t]
\small
\renewcommand{\arraystretch}{1.15}
    \centering
        \caption{Overall comparison on 10 real-world datasets. Performance ranks are shown in the parenthesis. Methods with the best performance are marked in bold. "OOM" refers to a Out of memory error with 256GB system RAM.
        }
        \vspace{-3mm}
    \setlength{\tabcolsep}{0.5mm}
    \resizebox{0.999\textwidth}{!}{
    \begin{tabular}{c|cccccccc|cccccccc}
    \hline
    \multirow{2}{*}{\textbf{Datasets}}  & \multicolumn{8}{c|}{\textbf{AUC-ROC}} & \multicolumn{8}{c}{\textbf{AUC-PR}}\\
    \cline{2-17}
  &\textbf{GAN} & \textbf{REPEN} & \textbf{DevNet} & \textbf{DeepSAD} & \textbf{FEAWAD} & \textbf{PReNet} & \textbf{SSIF} & \textbf{SetAD}  &\textbf{GAN} & \textbf{REPEN} & \textbf{DevNet} & \textbf{DeepSAD} & \textbf{FEAWAD} & \textbf{PReNet} & \textbf{SSIF} & \textbf{SetAD}\\
    \hline
    Cardio.
    &0.7308(8) &0.8930(5) &0.9332(2) &0.8827(6) &0.8975(4) &0.9325(3)&0.8759(7) &\textbf{0.9433(1)}
    &0.4457(8) &0.7483(6) &0.8314(3) &0.7505(5) &0.7984(4) &0.8331(2)&0.7377(7) &\textbf{0.8447(1)}\\
    Mammo.
    &0.8595(7) &0.8971(4) &0.9024(2) &0.8990(3) &0.8656(6) &0.8844(5)&0.6493(8) &\textbf{0.9032(1)}
    &0.1946(7) &0.4526(6) &0.5230(2) &0.5115(3) &0.4969(5) &0.5084(4) &0.1427(8) &\textbf{0.5650(1)}\\
    SpamBase 
    &0.8261(6) &0.8533(4) &0.8534(3) &0.7575(8) &0.8715(2) &0.8342(5) &0.8204(7) &\textbf{0.9185(1)}
    &0.7427(7) &0.7814(5) &0.8314(3) &0.6690(8) &0.8423(2) &0.8295(4) &0.7736(6) &\textbf{0.8806(1)}\\
    Mnist 
    &0.6995(8) &\textbf{0.9707(1)} &0.9361(5)          &0.9427(4) &0.9592(2) &0.9246(6) &0.8760(7) &0.9586(3)
    &0.2886(8) &0.8095(3)          &0.8331(2) &0.7387(6) &0.7670(5) &0.8038(4) &0.5120(7)& \textbf{0.8412(1)}\\
    Shuttle
    &0.9737(8) &0.9922(4) &0.9886(5) &\textbf{0.9966(1)} &0.9923(3) &0.9856(7) &0.9942(2) &0.9863(6)
    &0.9041(8) &0.9655(6) &\textbf{0.9770(1)} &0.9662(5) &0.9696(4) &0.9739(3) &0.9407(7) &0.9745(2)\\
    Celeba 
    &0.3773(7) &0.9268(4) &0.9372(2) &0.8630(6) &0.8640(5) &0.9327(3) &OOM &\textbf{0.9444(1)}
    &0.0218(7) &0.2201(4) &0.2508(2) &0.1467(6) &0.1906(5) &0.2496(3) &OOM  &\textbf{0.2653(1)}\\
    Cover
    &0.6257(7) &0.9960(5) &0.9995(2) &0.9929(6) &0.9961(4) &0.9994(3) &OOM &\textbf{0.9996(1)}
    &0.0225(7) &0.7688(6) &0.9547(2)          &0.8981(4) &0.8619(5) &0.9482(3) &OOM         &\textbf{0.9635(1)}\\
    Campaign 
    &0.6623(8) &0.8066(4) &0.8078(3) &0.7624(7) &0.7986(5) &0.7976(6) &0.8706(2) &\textbf{0.8813(1)}
    &0.2395(8) &0.4176(4) &0.4264(3) &0.3315(7) &0.3744(6) &0.4158(5) &0.4643(2) &\textbf{0.5158(1)}\\
    Fraud
    &0.8309(7) &0.9703(2) &0.9428(6) &0.9450(5) &0.9466(4) &0.9477(3) &OOM &\textbf{0.9710(1)}
    &0.1841(7) &0.6708(2) &0.5852(5) &0.6059(3) &0.5863(4) &0.5301(6) &OOM &\textbf{0.6855(1)}\\
    Census
    &0.6670(7) &\textbf{0.8995(1)} &0.8125(3) &0.7603(6) &0.8097(5) &0.8118(4) &OOM &0.8984(2)
    &0.0920(7) &0.4025(2) &0.4011(3) &0.2283(5) &0.2103(6) &0.3823(4) &OOM &\textbf{0.4719(1)}\\
    \hline
    Average
    &0.7253(7) &0.9205(2) &0.9113(3) &0.8802(6) &0.9001(5) &0.9051(4) &- &\textbf{0.9405(1)}
    &0.3136(7) &0.6237(4) &0.6614(2) &0.5846(6) &0.6098(5) &0.6475(3) &- &\textbf{0.7008(1)}\\
    
    \hline
    \end{tabular}
    }
    \vspace{-3mm}
    \label{tab2}
\end{table*}

\subsection{Experimental Setup}

\textit{\textbf{Datasets.}} We conduct extensive experiments on 10 real-world datasets from various domains, commonly used to benchmark anomaly detection models~\cite{han2022adbench, liu2008isolation, pang2019deep, zhou2021feature, pang2023deep}. These datasets are sourced from the ODDS library and the Adbench benchmark~\cite{han2022adbench}. We adopt standard anomaly definitions consistent with prior studies~\cite{bandaragoda2014efficient, aggarwal2015theoretical, liu2008isolation, han2022adbench}, leveraging domain-specific knowledge or identifying minority classes as anomalies. Specifically, in the Cardiotocography and Mammography datasets, which pertain to medical diagnosis, instances associated with specific diseases are labeled as anomalies. For SpamBase, which involves spam email detection, spam messages are regarded as anomalies. In the case of image-based datasets such as Shuttle, CelebA, and MNIST originating from handwritten digit recognition, spacecraft diagnostics, and facial attribute domains, the samples from infrequent classes are considered anomalous. The Campaign dataset, which records outcomes of bank telemarketing efforts, considers the rare successful cases as anomalies. For the Fraud dataset, which focuses on credit card transactions, fraudulent activities are treated as anomalies. Lastly, in the Census dataset derived from U.S. census data, individuals with unusually high incomes are identified as anomalies.

For each dataset, we randomly partition the data into an 80\% training set and a 20\% test set. Following existing methods, the training set is composed of a large pool of unlabeled data and a small, designated set of labeled anomalies, and the contamination rate within the unlabeled data is maintained under 2\%~\cite{pang2019deep, zhou2021feature, pang2023deep}. The number of labeled anomalies provided for supervision is kept small and proportional to the dataset's scale to reflect practical constraints where such labels are scarce. This ensures a consistent, limited amount of supervision across all datasets. The specific details of each dataset are provided in Table \ref{tab1}. Prior to training, all datasets undergo z-score normalization. Any feature dimensions with a standard deviation of zero are removed as they provide no discriminative information. For evaluation on the test set, we select the model checkpoint that achieves the lowest loss value on the training data.

\textit{\textbf{Evaluation Metrics.}} To provide a comprehensive and robust assessment of model performance, we employ two standard metrics: the Area Under the Receiver Operating Characteristic Curve (AUC-ROC) and the Area Under the Precision-Recall Curve (AUC-PR). AUC-ROC provides a general measure of a model's ability to discriminate between classes across all decision thresholds. However, in anomaly detection scenarios characterized by extreme class imbalance, AUC-ROC can be overly optimistic, as high scores can be achieved by correctly classifying the abundant normal instances, even if performance on the rare anomaly class is poor. To provide a more focused assessment on the minority class, we therefore also report the AUC-PR. This metric evaluates the trade-off between precision and recall and is widely recognized as a more informative measure when the positive class (anomalies) is rare. This is because its value is not influenced by the number of true negative classifications, which dominate imbalanced datasets. For both metrics, a higher value indicates superior performance..

\textit{\textbf{Competing Methods.}}
We benchmark SetAD against six state-of-the-art SSAD models to ensure a comprehensive evaluation: REPEN\cite{pang2018learning}, DeepSAD~\cite{DeepSAD}, GANormaly~\cite{akcay2019ganomaly}, DevNet\cite{pang2019deep}, FEAWAD~\cite{zhou2021feature}, PReNet\cite{pang2023deep} and SSIF~\cite{stradiotti2024semi}. This selection covers a wide spectrum of modeling philosophies. GANormaly represents generative approaches that leverage adversarial training and reconstruction-based approaches. DevNet, FEAWAD, and the Isolation Forest-based SSIF represent point-wise approaches that score individual instances. DeepSAD and PReNet exemplify center-based and pairwise methods, respectively, evaluating a point based on its distance to a learned center or its relationship with another point. Finally, REPEN serves as a simplified set-based baseline, scoring a point via its k-NN distance to a reference set of normals. 
Comparing with these diverse methods helps validate the benefits of our expressive, interaction-aware set-based approach.

\textit{\textbf{Implementation Details.}}
For our model, the size of the set $k$ is set to 8 in default. We randomly sample sets containing 0, 1 and 2 anomalies for training in the graded learning stage. The embedding function $\mathcal{E}$ is implemented as a single-layer MLP with a ReLU activation, which projects each input point into a 20-dimensional latent space, a dimension chosen in line with prior works~\cite{pang2019deep, pang2018learning, pang2023deep}. The self-attention mechanism utilizes 2 attention heads. We experimented with a higher number of heads (e.g., 4 and 5) but observed no significant performance gain, thus selecting 2 for model simplicity. The regression head $\mathcal{R}$ is also a single-layer MLP that maps the aggregated set representation to a final scalar score. The model is trained for 20 epochs, with each epoch consisting of 20 mini-batches. We use the RMSProp optimizer~\cite{zou2019sufficient} with a learning rate of 1e-3 and a weight decay of 0.1 to update the model parameters. 
In context-calibrated anomaly scoring, we vary \(n_r\) from 10 to 50 and observe that performance stabilizes around 30. 
For \(n_C\), performance improves steadily from 10 to 80, and we choose \(n_C = 60\) as a practical trade-off between accuracy and computational cost.
For all competing methods, we use their officially released code and follow the hyperparameter settings recommended in their original papers. To ensure robust and reproducible results, we report the average performance over ten independent runs for all models. 

 
\begin{figure*}[t!]
\centering
\vspace{-1ex}
{
\includegraphics[width = 0.95\linewidth]{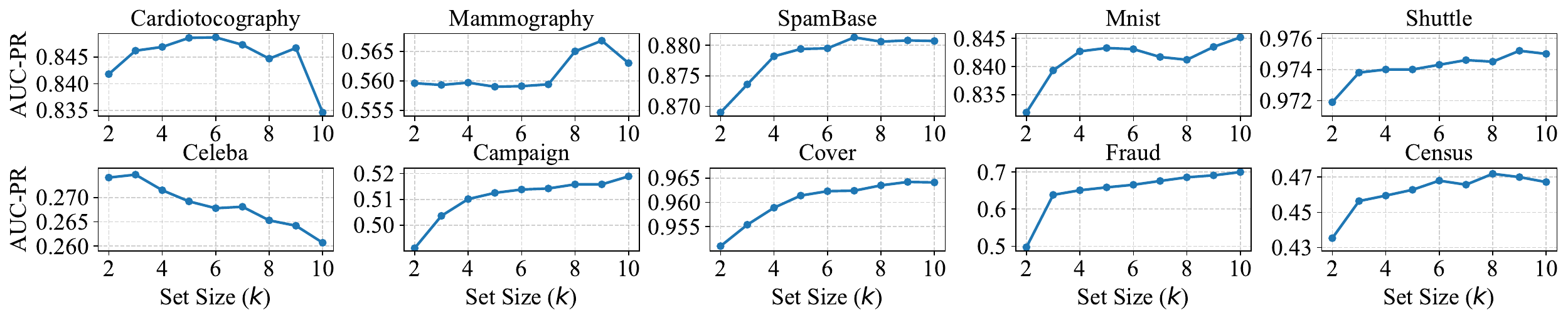}
}
\vspace{-3ex}
\caption{AUC-PR w.r.t different set size $k$.}
\vspace{-4mm}
\label{fig3}
\end{figure*}

\subsection{Overall Performance} \label{sec4.3}
We evaluate the overall effectiveness of SetAD by comparing it against seven state-of-the-art methods on 10 real-world datasets. The detailed results for both AUC-ROC and AUC-PR are presented in Table \ref{tab2}. Note that among the baselines, SSIF can not operate with 256GB system RAM on 4 large datasets. The experimental results clearly demonstrate the superiority of our proposed SetAD framework. As shown in the table, SetAD achieves the best average performance across all datasets on both metrics, ranking first in 7 out of 10 datasets on AUC-ROC and a remarkable 9 out of 10 on AUC-PR. In detail, SetAD obtains better average AUC-ROC score than REPEN (2.2\%), DevNet (3.2\%), PReNet (3.9\%), FEAWAD (4.5\%), DeepSAD (6.9\%), and GAN (29.7\%). For AUC-PR, SetAD achieves more substantial improvements compared to DevNet (6.0\%), PReNet (8.2\%), REPEN (12.4\%), FEAWAD (14.9\%), DeepSAD (19.9\%), and GAN (123.5\%). These results demonstrate the effectiveness of our proposed approach across various settings.

This strong performance can be attributed to the core design principles of SetAD. Unlike point-wise methods like DevNet and FEAWAD, or pairwise models like PReNet, which rely on localized comparisons, SetAD's self-attention based encoder is designed to capture the holistic, high-order interactions within a set of points. This allows it to learn more discriminative representations by understanding how an anomaly disrupts the collective structure of its local context. When compared to REPEN which also considers a set of normals, SetAD's ability to model complex interactions proves superior to REPEN's simpler aggregation of k-NN distances.
Notably, SetAD's advantage is pronounced in the AUC-PR metric, which is more sensitive to performance on the rare anomaly class. This suggests that by learning from the graded anomalousness of entire sets, our model develops a more precise understanding of the decision boundary, leading to fewer false positives and a more practical utility in real-world, imbalanced scenarios. These comprehensive results provide strong empirical validation for our central hypothesis: reframing anomaly detection as a set-level learning task leads to more effective models.

\vspace{-0.5em}
\subsection{Further Analysis}

\paragraph{\textbf{Impact of Set Size ($k$).}} The core argument of this paper is that evaluating a data point within a richer context enables a more robust and accurate assessment of its anomalousness. 
To validate this, we systematically vary the set size $k$ from 2 to 10 and evaluate SetAD's performance using AUC-PR across different datasets. Figure~\ref{fig3} presents per-dataset results to clearly show trends despite scale differences. As shown, 
increasing the set size beyond a simple pair ($k=2$) leads to a significant performance   on nearly all datasets. For instance, on datasets like Fraud and Campaign, performance continues to rise even as the set size approaches 10, suggesting that these datasets contain complex anomalies whose characteristics are best revealed through a larger contextual window. On other datasets such as Mnist and Census, performance peaks at an intermediate set size before plateauing, indicating that a moderately sized context is sufficient to capture the necessary discriminative information. The only exception is the Celeba dataset, where performance peaks at $k=3$ before gradually declining. This behavior could be attributed to several factors, such as a data structure where distant points offer little relevant information and may even obscure the signal from closer neighbors. Nevertheless, the optimal performance is still achieved by moving beyond a simple pair. While the ideal context size may vary on different datasets, the optimal performance is achieved at a size strictly greater than 2. This indicates the necessity of a more expressive set-based approach. 

In summary, these findings robustly support our core claim. By designing an architecture capable of modeling high-order interactions, SetAD translates larger contexts into better performance. Based on these results, which show strong and stable performance around a size of 8 across many datasets, we set $k=8$ as the default set size for all other experiments in this paper. While there remains room for future work in areas like adaptive context selection or more advanced set encoders, this analysis shows advantage of the set-based formulation for semi-supervised anomaly detection.

\begin{figure}[t]
\centering
    {
    \includegraphics[width=0.35\textwidth, trim=0 0 0 30, clip]{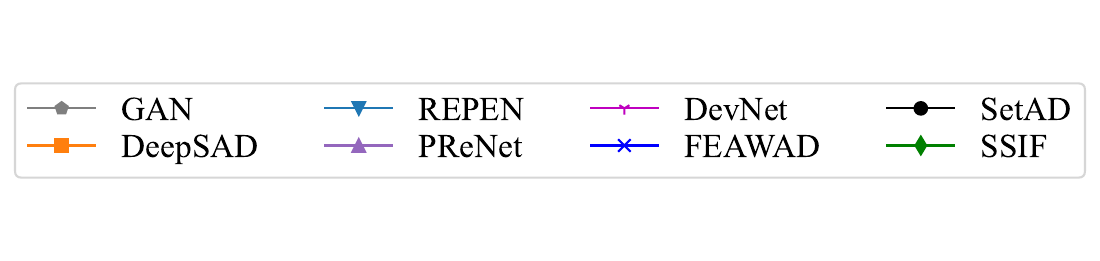}
    }
\\
\vspace{-3ex}
{
\includegraphics[height = 0.27\textwidth]{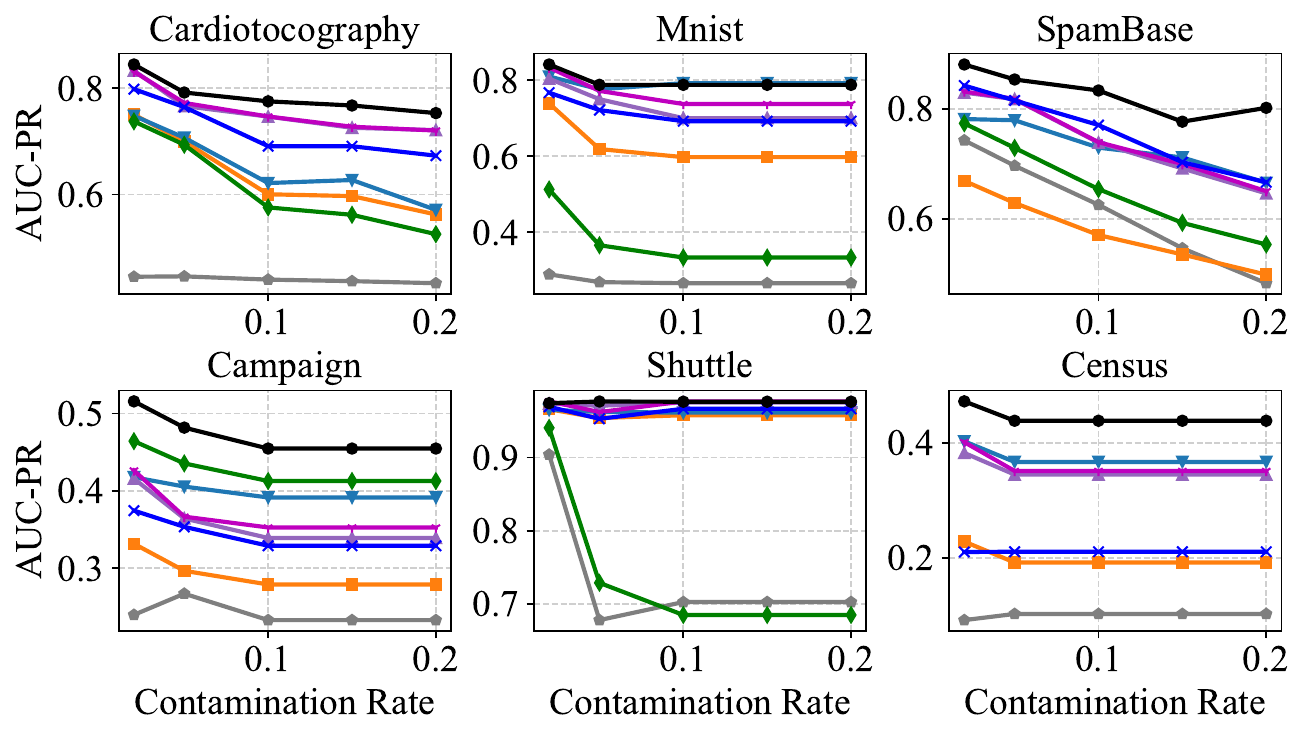}
}
\vspace{-5ex}
\caption{AUC-PR w.r.t different contamination rates.}
\vspace{-2.5ex}
\label{fig2}
\end{figure}

\textit{\textbf{Robustness to Anomaly Contamination.}}
A critical challenge for semi-supervised AD models is their ability to perform reliably when the unlabeled training data is contaminated with a significant number of unknown anomalies. To evaluate SetAD's resilience to this, we conduct an experiment on six datasets where the natural anomaly contamination rate exceeds 2\%. We systematically increase the contamination rate of the unlabeled training data from 2\% up to 20\% and plot the resulting AUC-PR performance for all competing methods. The trends are illustrated in Figure \ref{fig2}. Note that for some datasets like Mnist and Shuttle, the performance of most models stabilizes after a certain point, which occurs because the dataset's natural contamination rate is reached before hitting the 20\% mark.

From the experimental results we can see that SetAD consistently maintains the highest performance across nearly all contamination levels and datasets. As shown in the plots for Cardiotocography, SpamBase, and Campaign, even as the training data becomes increasingly noisy, SetAD's performance remains superior to all baselines, demonstrating its strong effectiveness. Additionally, SetAD exhibits a more graceful performance degradation compared to most competing methods. While all models inevitably suffer as the contamination rate rises, SetAD's performance curve tends to be flatter, indicating a higher degree of robustness. This phenomenon is particularly evident in the SpamBase dataset. While the performance of most competing methods declines sharply with rising contamination, SetAD's performance curve remains relatively flat. This enhanced resilience stems directly from the core principles of our design. On one hand, the set-based evaluation, which leverages a richer context to provide more stable signal for identifying anomalies. On the other hand, our context-calibrated anomaly scoring strategy is explicitly designed to counteract contamination. By normalizing a point's score against its peers within the same context, it actively filters out the systemic bias introduced by unknown anomalies in the training pool. Overall, this analysis confirms that SetAD not only performances well under clean conditions but also more robust for real-world scenarios with more data contamination.

\begin{table}[t]
\centering
\small
\caption{Ablation study on the inference strategy.}
\vspace{-3mm}
\resizebox{0.38\textwidth}{!}{
\begin{tabular}{lcccc}
\toprule
\multirow{2}{*}{\textbf{Dataset}} & \multicolumn{2}{c}{\textbf{AUC-ROC}} & \multicolumn{2}{c}{\textbf{AUC-PR}} \\
\cmidrule(lr){2-3} \cmidrule(lr){4-5}
& \textbf{w/o} CAS & SetAD & \textbf{w/o} CAS & SetAD \\
\midrule
Cardiotocography & \textbf{0.9454} & 0.9433 & \textbf{0.8480} & 0.8447 \\
Mammography      & 0.8986 & \textbf{0.9032} & 0.5539 & \textbf{0.5650} \\
SpamBase         & 0.9098 & \textbf{0.9185} & 0.8745 & \textbf{0.8806} \\
Mnist            & 0.9564 & \textbf{0.9586} & 0.8350 & \textbf{0.8412} \\
Shuttle          & 0.9855 & \textbf{0.9863} & 0.9721 & \textbf{0.9745} \\
Celeba           & 0.9373 & \textbf{0.9444} & 0.2541 & \textbf{0.2653} \\
Cover            & 0.9994 & \textbf{0.9996} & 0.9570 & \textbf{0.9635} \\
Campaign         & 0.8674 & \textbf{0.8813} & 0.5064 & \textbf{0.5158} \\
Fraud            & 0.9686 & \textbf{0.9710} & \textbf{0.6860} & 0.6855 \\
Census           & 0.8843 & \textbf{0.8984} & 0.4556 & \textbf{0.4719} \\
\midrule
\textbf{AVG}     & 0.9353 & \textbf{0.9405} & 0.6943 & \textbf{0.7008} \\
\bottomrule
\end{tabular}
}
\vspace{-3mm}
\label{ablation}
\end{table}

\textit{\textbf{Effectiveness of Context-Calibrated Anomaly Scoring.}} 
To verify the contribution of our proposed inference strategy, we conduct an ablation study comparing the full SetAD model with a variant that omits the context-calibrated anomaly scoring (CAS) step. For this variant, denoted as "w/o CAS", a point's anomaly score is derived from the raw output of the set encoder, averaged over multiple random contexts without peer-based normalization. Table \ref{ablation} shows that the full SetAD model consistently outperforms the variant without CAS on the vast majority of datasets, particularly in terms of the more challenging AUC-PR metric. On average, the inclusion of CAS brings a notable improvement, demonstrating its overall effectiveness. By normalizing a test point's score against the baseline established by its normal peers within the same context, this strategy effectively cancels out the scoring bias that could be introduced by a "dirty" or unrepresentative contexts. Interestingly, on a few datasets like Cardiotocography, the performance is very close, and the variant occasionally performs slightly better. This suggests that the raw output of the set encoder may already be sufficient for effective ranking. However, the strong average performance gain confirms that context-calibrated anomaly scoring is a necessary component for ensuring the robustness of SetAD across diverse and potentially noisy real-world scenarios.


\label{Efficiency}
\begin{figure}[t!]
\centering
    {
    \includegraphics[width=0.35\textwidth, trim=0 0 0 30, clip]{./images/legend_only.pdf}
    }
\\
\vspace{-3ex}
{
\includegraphics[width = 0.49\textwidth, trim=7 0 0 0, clip]{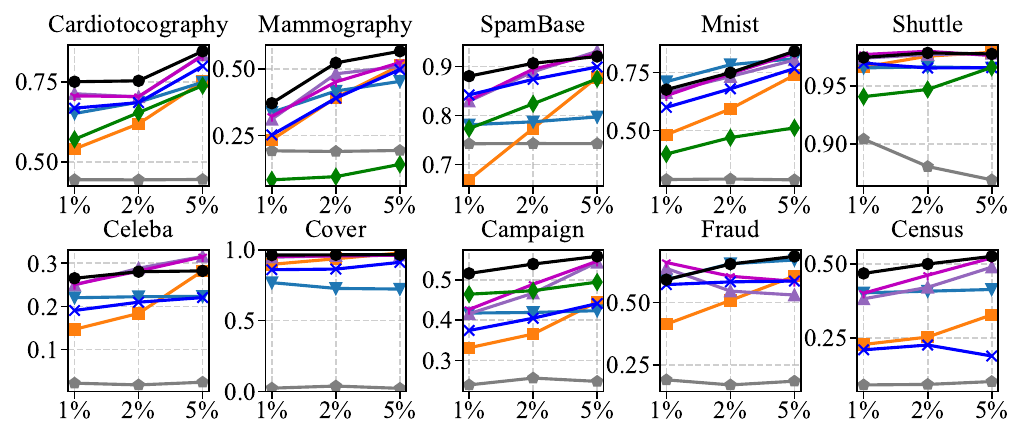}
}
\vspace{-4.5ex}
\caption{AUC-PR w.r.t labeled ratio.}
\vspace{-3ex}
\label{fig4}
\end{figure}

\textit{\textbf{Data Efficiency.}} To comprehensively evaluate the robustness and data efficiency of our proposed SetAD, we conducted a sensitivity analysis on the proportion of available labeled anomalies, as a model's utility in real-world scenarios often depends on its performance with scarce supervision. In this experiment, we varied the ratio of labeled anomalies used for training, selecting 1\%, 2\%, and 5\% of the total available anomalies from each dataset's training split. We benchmarked all competing models under these three settings, using the Area Under the Precision-Recall Curve (AUC-PR) as the primary performance metric due to its suitability for imbalanced datasets. The detailed performance trends for all models across the ten datasets are visualized in Figure \ref{fig4}.
The results reveal several key insights. Firstly, most semi-supervised deep learning models, including DeepSAD, PReNet, DevNet, and our SetAD, generally exhibit improved performance as the ratio of labeled anomalies increases, underscoring the value of supervision. Secondly, and most importantly, SetAD consistently outperforms all baselines across nearly all datasets and at all label ratios. This performance gap is often most pronounced in the most challenging low-label setting (1\%). For example, on datasets like Campaign, SetAD's score at 1\% already surpasses that of many competing methods at 5\%, demonstrating the superior data efficiency of our set-level learning approach. The combinatorial nature of our training allows the model to extract a rich supervisory signal even from very few labeled examples. Finally, SetAD demonstrates highly stable and robust learning, with its performance curve showing consistent improvement as more labels are introduced. In summary, this analysis validates that SetAD not only achieves state-of-the-art performance but also exhibits remarkable robustness and efficiency, especially when labeled data is extremely limited.

\section{Conclusion}
In this paper, we revisited the foundational principle that anomalies are defined by their deviation from a collective group. We argued that many existing semi-supervised methods focus on individual points or simple pairs, which makes them fail to fully capture the rich, high-order context required to detect complex anomalies. To address this gap, we introduced SetAD, a novel framework that reframes anomaly detection as a set-level graded learning task. Our approach trains an interaction-aware set encoder, based on a self-attention mechanism, to quantify the degree of anomalousness within a set of points. This is complemented by a robust, context-calibrated anomaly scoring strategy that provides stable and well-calibrated anomaly scores. 
Our experiments not only demonstrate that SetAD significantly outperforms state-of-the-art methods but also show that its performance consistently improves with increased context size—a crucial finding that highlights the tangible benefits of a holistic, set-based approach.


\bibliographystyle{ACM-Reference-Format}
\balance
\bibliography{references}

@inproceedings{li2020copod,
  title={COPOD: copula-based outlier detection},
  author={Li, Zheng and Zhao, Yue and Botta, Nicola and Ionescu, Cezar and Hu, Xiyang},
  booktitle={2020 IEEE international conference on data mining (ICDM)},
  pages={1118--1123},
  year={2020},
  organization={IEEE}
}

@inproceedings{zou2019sufficient,
  title={A sufficient condition for convergences of adam and rmsprop},
  author={Zou, Fangyu and Shen, Li and Jie, Zequn and Zhang, Weizhong and Liu, Wei},
  booktitle={Proceedings of the IEEE/CVF Conference on computer vision and pattern recognition},
  pages={11127--11135},
  year={2019}
}

@article{li2022ecod,
  title={Ecod: Unsupervised outlier detection using empirical cumulative distribution functions},
  author={Li, Zheng and Zhao, Yue and Hu, Xiyang and Botta, Nicola and Ionescu, Cezar and Chen, George H},
  journal={IEEE Transactions on Knowledge and Data Engineering},
  volume={35},
  number={12},
  pages={12181--12193},
  year={2022},
  publisher={IEEE}
}

@inproceedings{pang2019deep,
  title={Deep anomaly detection with deviation networks},
  author={Pang, Guansong and Shen, Chunhua and Van Den Hengel, Anton},
  booktitle={Proceedings of the 25th ACM SIGKDD international conference on knowledge discovery \& data mining},
  pages={353--362},
  year={2019}
}

@inproceedings{chen2017outlier,
  title={Outlier detection with autoencoder ensembles},
  author={Chen, Jinghui and Sathe, Saket and Aggarwal, Charu and Turaga, Deepak},
  booktitle={Proceedings of the 2017 SIAM international conference on data mining},
  pages={90--98},
  year={2017},
  organization={SIAM}
}

@inproceedings{zhou2017anomaly,
  title={Anomaly detection with robust deep autoencoders},
  author={Zhou, Chong and Paffenroth, Randy C},
  booktitle={Proceedings of the 23rd ACM SIGKDD international conference on knowledge discovery and data mining},
  pages={665--674},
  year={2017}
}

@article{zhou2021feature,
  title={Feature encoding with autoencoders for weakly supervised anomaly detection},
  author={Zhou, Yingjie and Song, Xucheng and Zhang, Yanru and Liu, Fanxing and Zhu, Ce and Liu, Lingqiao},
  journal={IEEE Transactions on Neural Networks and Learning Systems},
  volume={33},
  number={6},
  pages={2454--2465},
  year={2021},
  publisher={IEEE}
}

@inproceedings{schlegl2017unsupervised,
  title={Unsupervised anomaly detection with generative adversarial networks to guide marker discovery},
  author={Schlegl, Thomas and Seeb{\"o}ck, Philipp and Waldstein, Sebastian M and Schmidt-Erfurth, Ursula and Langs, Georg},
  booktitle={International conference on information processing in medical imaging},
  pages={146--157},
  year={2017},
  organization={Springer}
}

@inproceedings{breunig2000lof,
  title={LOF: identifying density-based local outliers},
  author={Breunig, Markus M and Kriegel, Hans-Peter and Ng, Raymond T and Sander, J{\"o}rg},
  booktitle={Proceedings of the 2000 ACM SIGMOD international conference on Management of data},
  pages={93--104},
  year={2000}
}

@article{scholkopf2001estimating,
  title={Estimating the support of a high-dimensional distribution},
  author={Sch{\"o}lkopf, Bernhard and Platt, John C and Shawe-Taylor, John and Smola, Alex J and Williamson, Robert C},
  journal={Neural computation},
  volume={13},
  number={7},
  pages={1443--1471},
  year={2001},
  publisher={MIT Press One Rogers Street, Cambridge, MA 02142-1209, USA journals-info~…}
}

@inproceedings{ruff2018deep,
  title={Deep one-class classification},
  author={Ruff, Lukas and Vandermeulen, Robert and Goernitz, Nico and Deecke, Lucas and Siddiqui, Shoaib Ahmed and Binder, Alexander and M{\"u}ller, Emmanuel and Kloft, Marius},
  booktitle={International conference on machine learning},
  pages={4393--4402},
  year={2018},
  organization={PMLR}
}

@inproceedings{pang2023deep,
  title={Deep weakly-supervised anomaly detection},
  author={Pang, Guansong and Shen, Chunhua and Jin, Huidong and van den Hengel, Anton},
  booktitle={Proceedings of the 29th ACM SIGKDD Conference on Knowledge Discovery and Data Mining},
  pages={1795--1807},
  year={2023}
}

@inproceedings{pang2018learning,
  title={Learning representations of ultrahigh-dimensional data for random distance-based outlier detection},
  author={Pang, Guansong and Cao, Longbing and Chen, Ling and Liu, Huan},
  booktitle={Proceedings of the 24th ACM SIGKDD international conference on knowledge discovery \& data mining},
  pages={2041--2050},
  year={2018}
}

@article{han2022adbench,
  title={Adbench: Anomaly detection benchmark},
  author={Han, Songqiao and Hu, Xiyang and Huang, Hailiang and Jiang, Minqi and Zhao, Yue},
  journal={Advances in Neural Information Processing Systems},
  volume={35},
  pages={32142--32159},
  year={2022}
}

@inproceedings{akcay2019ganomaly,
  title={Ganomaly: Semi-supervised anomaly detection via adversarial training},
  author={Akcay, Samet and Atapour-Abarghouei, Amir and Breckon, Toby P},
  booktitle={Computer Vision--ACCV 2018: 14th Asian Conference on Computer Vision, Perth, Australia, December 2--6, 2018, Revised Selected Papers, Part III 14},
  pages={622--637},
  year={2019},
  organization={Springer}
}

@inproceedings{DeepSAD,
  author       = {Lukas Ruff and
                  Robert A. Vandermeulen and
                  Nico G{\"{o}}rnitz and
                  Alexander Binder and
                  Emmanuel M{\"{u}}ller and
                  Klaus{-}Robert M{\"{u}}ller and
                  Marius Kloft},
  title        = {Deep Semi-Supervised Anomaly Detection},
  booktitle    = {8th International Conference on Learning Representations, {ICLR}},
  year         = {2020},
}

@inproceedings{liu2008isolation,
  title={Isolation forest},
  author={Liu, Fei Tony and Ting, Kai Ming and Zhou, Zhi-Hua},
  booktitle={2008 eighth ieee international conference on data mining},
  pages={413--422},
  year={2008},
  organization={IEEE}
}

@article{grubbs1969procedures,
  title={Procedures for detecting outlying observations in samples},
  author={Grubbs, Frank E},
  journal={Technometrics},
  volume={11},
  number={1},
  pages={1--21},
  year={1969},
  publisher={Taylor \& Francis}
}

@inproceedings{liu2021pick,
  title={Pick and choose: a GNN-based imbalanced learning approach for fraud detection},
  author={Liu, Yang and Ao, Xiang and Qin, Zidi and Chi, Jianfeng and Feng, Jinghua and Yang, Hao and He, Qing},
  booktitle={Proceedings of the web conference 2021},
  pages={3168--3177},
  year={2021}
}

@article{dal2017credit,
  title={Credit card fraud detection: a realistic modeling and a novel learning strategy},
  author={Dal Pozzolo, Andrea and Boracchi, Giacomo and Caelen, Olivier and Alippi, Cesare and Bontempi, Gianluca},
  journal={IEEE transactions on neural networks and learning systems},
  volume={29},
  number={8},
  pages={3784--3797},
  year={2017},
  publisher={IEEE}
}

@article{garcia2009anomaly,
  title={Anomaly-based network intrusion detection: Techniques, systems and challenges},
  author={Garcia-Teodoro, Pedro and Diaz-Verdejo, Jesus and Maci{\'a}-Fern{\'a}ndez, Gabriel and V{\'a}zquez, Enrique},
  journal={computers \& security},
  volume={28},
  number={1-2},
  pages={18--28},
  year={2009},
  publisher={Elsevier}
}

@article{henson2021anomaly,
  title={Anomaly detection to predict relapse risk in schizophrenia},
  author={Henson, Philip and D’Mello, Ryan and Vaidyam, Aditya and Keshavan, Matcheri and Torous, John},
  journal={Translational psychiatry},
  volume={11},
  number={1},
  pages={28},
  year={2021},
  publisher={Nature Publishing Group UK London}
}

@article{fernando2021deep,
  title={Deep learning for medical anomaly detection--a survey},
  author={Fernando, Tharindu and Gammulle, Harshala and Denman, Simon and Sridharan, Sridha and Fookes, Clinton},
  journal={ACM Computing Surveys (CSUR)},
  volume={54},
  number={7},
  pages={1--37},
  year={2021},
  publisher={ACM New York, NY, USA}
}

@article{zenati2018efficient,
  title={Efficient gan-based anomaly detection},
  author={Zenati, Houssam and Foo, Chuan Sheng and Lecouat, Bruno and Manek, Gaurav and Chandrasekhar, Vijay Ramaseshan},
  journal={arXiv preprint arXiv:1802.06222},
  year={2018}
}

@inproceedings{bandaragoda2014efficient,
  title={Efficient anomaly detection by isolation using nearest neighbour ensemble},
  author={Bandaragoda, Tharindu R and Ting, Kai Ming and Albrecht, David and Liu, Fei Tony and Wells, Jonathan R},
  booktitle={2014 IEEE International conference on data mining workshop},
  pages={698--705},
  year={2014},
  organization={IEEE}
}

@article{aggarwal2015theoretical,
  title={Theoretical foundations and algorithms for outlier ensembles},
  author={Aggarwal, Charu C and Sathe, Saket},
  journal={Acm sigkdd explorations newsletter},
  volume={17},
  number={1},
  pages={24--47},
  year={2015},
  publisher={ACM New York, NY, USA}
}

@inproceedings{gao2025semi,
  title={Semi-Supervised Anomaly Detection through Denoising-Aware Contrastive Distance Learning},
  author={Gao, Jianling and Tao, Chongyang and Sun, Zhenchao and Jiang, Xiya and Ma, Shuai},
  booktitle={Proceedings of the ACM on Web Conference 2025},
  pages={2111--2119},
  year={2025}
}

@inproceedings{stradiotti2024semi,
  title={Semi-supervised isolation forest for anomaly detection},
  author={Stradiotti, Luca and Perini, Lorenzo and Davis, Jesse},
  booktitle={Proceedings of the 2024 SIAM International Conference on Data Mining (SDM)},
  pages={670--678},
  year={2024},
  organization={SIAM}
}

@article{xu2023deep,
  title={Deep isolation forest for anomaly detection},
  author={Xu, Hongzuo and Pang, Guansong and Wang, Yijie and Wang, Yongjun},
  journal={IEEE Transactions on Knowledge and Data Engineering},
  volume={35},
  number={12},
  pages={12591--12604},
  year={2023},
  publisher={IEEE}
}

@inproceedings{fu2024dense,
  title={Dense projection for anomaly detection},
  author={Fu, Dazhi and Zhang, Zhao and Fan, Jicong},
  booktitle={Proceedings of the AAAI conference on artificial intelligence},
  volume={38},
  number={8},
  pages={8398--8408},
  year={2024}
}

@article{xiao2025semi,
  title={Semi-Supervised Anomaly Detection Using Restricted Distribution Transformation},
  author={Xiao, Feng and Wang, Youqing and Qin, S Joe and Fan, Jicong},
  journal={IEEE Transactions on Neural Networks and Learning Systems},
  year={2025},
  publisher={IEEE}
}

@inproceedings{lu2024robust,
  title={A robust prioritized anomaly detection when not all anomalies are of primary interest},
  author={Lu, Guanyu and Zhou, Fang and Pavlovski, Martin and Zhou, Chenyi and Jin, Cheqing},
  booktitle={2024 IEEE 40th International Conference on Data Engineering (ICDE)},
  pages={775--788},
  year={2024},
  organization={IEEE}
}

@article{dong2024nng,
  title={Nng-mix: Improving semi-supervised anomaly detection with pseudo-anomaly generation},
  author={Dong, Hao and Frusque, Ga{\"e}tan and Zhao, Yue and Chatzi, Eleni and Fink, Olga},
  journal={IEEE Transactions on Neural Networks and Learning Systems},
  year={2024},
  publisher={IEEE}
}

@inproceedings{zhao2024weakly,
  title={Weakly supervised anomaly detection via knowledge-data alignment},
  author={Zhao, Haihong and Zi, Chenyi and Liu, Yang and Zhang, Chen and Zhou, Yan and Li, Jia},
  booktitle={Proceedings of the ACM Web Conference 2024},
  pages={4083--4094},
  year={2024}
}

@article{liu2024deep,
  title={Deep industrial image anomaly detection: A survey},
  author={Liu, Jiaqi and Xie, Guoyang and Wang, Jinbao and Li, Shangnian and Wang, Chengjie and Zheng, Feng and Jin, Yaochu},
  journal={Machine Intelligence Research},
  volume={21},
  number={1},
  pages={104--135},
  year={2024},
  publisher={Springer}
}

@article{vaswani2017attention,
  title={Attention is all you need},
  author={Vaswani, Ashish and Shazeer, Noam and Parmar, Niki and Uszkoreit, Jakob and Jones, Llion and Gomez, Aidan N and Kaiser, {\L}ukasz and Polosukhin, Illia},
  journal={Advances in neural information processing systems},
  volume={30},
  year={2017}
}

\appendix
\newpage

\end{document}